\documentclass[letterpaper, 10 pt, conference]{ieeeconf}  

\usepackage[T1]{fontenc}
\usepackage{cite}
\usepackage{amssymb,amsfonts}
\usepackage{blindtext}
\makeatletter
\let\NAT@parse\undefined
\makeatother
\usepackage{hyperref}
\usepackage{algorithmic}
\usepackage{graphicx}
\usepackage{textcomp}
\usepackage{mathtools}
\usepackage{changes}
\usepackage[font=footnotesize,labelformat=simple]{subcaption}
\usepackage{algorithm}
\usepackage{xcolor}
\usepackage{color, soul}
\usepackage{amsmath}
\usepackage{booktabs}
\usepackage{multirow}
\usepackage{xstring}
\usepackage{hhline}
\usepackage{kotex}
\usepackage{siunitx}
\usepackage{comment}
\usepackage{physics}
\usepackage{tikz}
\usepackage{cleveref}
\usepackage{gensymb}
\usepackage{threeparttable}
\usepackage{caption}
\usepackage{url_bib_utils}

\crefformat{section}{\S#2#1#3} 
\crefformat{subsection}{\S#2#1#3}
\crefformat{subsubsection}{\S#2#1#3}

\definecolor{pr}{RGB}{0, 0, 0}
\definecolor{gl}{RGB}{0, 0, 0}
\definecolor{v3}{RGB}{0,0,0}
\definecolor{v5}{RGB}{0,0,0}
\definecolor{v7}{RGB}{0,0,255}
\definecolor{greenfoot}{RGB}{126,170,85}
\definecolor{orangefoot}{RGB}{222,131,68}

\IEEEoverridecommandlockouts                              

\title{\LARGE \bf
Robust Recovery Motion Control for Quadrupedal Robots\\ via Learned Terrain Imagination
}

\author{I Made Aswin Nahrendra, Minho Oh, Byeongho Yu, Hyungtae Lim, and Hyun Myung$^{*}$, \textit{Senior Member, IEEE}
\thanks{This work was supported by Korea Evaluation Institute of Industrial Technology (KEIT) funded by the Korea Government (MOTIE) under grant No. 20019216, ``Development of Mobile Intelligence SW for Autonomous Navigation of Legged Robots in Dynamic and Atypical Environments for Real Application''. The students are supported by BK21 FOUR.}
\thanks{The authors are with the School of Electrical Engineering at Korea Advanced Institute of Science and Technology (KAIST), Daejeon, 34141, Republic of Korea.  
{\tt\footnotesize \{anahrendra, minho.oh, bhyu, shapelim, hmyung\}@kaist.ac.kr}\hfill \break
\indent{$^*$Corresponding author: Hyun Myung}}
}

\begin{document}

\maketitle



\IEEEpeerreviewmaketitle

\begin{abstract}
Quadrupedal robots have emerged as a cutting--edge platform for assisting humans, finding applications in tasks related to inspection and exploration in remote areas. Nevertheless, their floating base structure renders them susceptible to fall in cluttered environments, where manual recovery by a human operator may not always be feasible. Several recent studies have presented recovery controllers employing deep reinforcement learning algorithms. However, these controllers are not specifically designed to operate effectively in cluttered environments, such as stairs and slopes, which restricts their applicability. In this study, we propose a robust all-terrain recovery policy to facilitate rapid and secure recovery in cluttered environments. We substantiate the superiority of our proposed approach through simulations and real-world tests encompassing various terrain types.
\end{abstract}

\section{Introduction}
\noindent In recent years, quadrupedal robot research has made significant strides in enhancing the mobility of ground mobile robots across challenging terrains~\cite{lee2020learning,miki2022learning,nahrendra2023dreamwaq}. Inspired by their animal counterparts, these robots possess the capability to navigate diverse terrain types and accomplish a wide range of tasks. The advent of deep reinforcement learning (RL) techniques has played a pivotal role in augmenting the agility and resilience of quadrupedal robots in natural, unstructured environments~\cite{lee2020learning,miki2022learning,nahrendra2023dreamwaq,kumar2021rma,fu2021minimizing}.

Despite the remarkable progress made in enhancing the robustness of quadrupedal robots\cite{lee2020learning,miki2022learning,nahrendra2023dreamwaq,kumar2021rma}, their performance in field operations is still susceptible to fall due to the environment's unique characteristics. Achieving a perfect success rate of $100\,\%$ in highly cluttered environments remains challenging. Similarly, real animals also face unexpected falls, highlighting the inherent difficulty of maintaining stability on four legs. However, animals can learn to swiftly recover from failure states through their experiences. Consequently, the operation of quadrupedal robots in natural environments necessitates the development of a robust and expeditious recovery strategy to ensure uninterrupted functionality.

To the best of our knowledge, the initial implementation of a robust recovery controller using a learning framework was introduced by~Lee~\textit{et al.}~\cite{lee2019robust}. While their work introduced a new approach to designing a resilient recovery controller, it relied on a complex hierarchical framework that segregated the self-righting and standing-up behaviors into separate policies. Additionally, their method was solely tested in a controlled laboratory setting on flat surfaces, limiting its evaluation to such conditions and thus not closely demonstrating generalized applicabilities in various cluttered environments.

A recent study developed a relatively straightforward recovery controller~\cite{smith2022legged} that serves as a reset mechanism for refining locomotion policies in the real--world. Notably, the recovery controller employs a single policy trained using a motion imitation framework. Nonetheless, akin to the limitations observed in~\cite{lee2019robust}, this recovery controller is primarily effective on relatively flat surfaces.

\begin{figure}[!t]
	\centering 
	\begin{subfigure}[b]{0.48\textwidth}
		\includegraphics[width=1.0\textwidth]{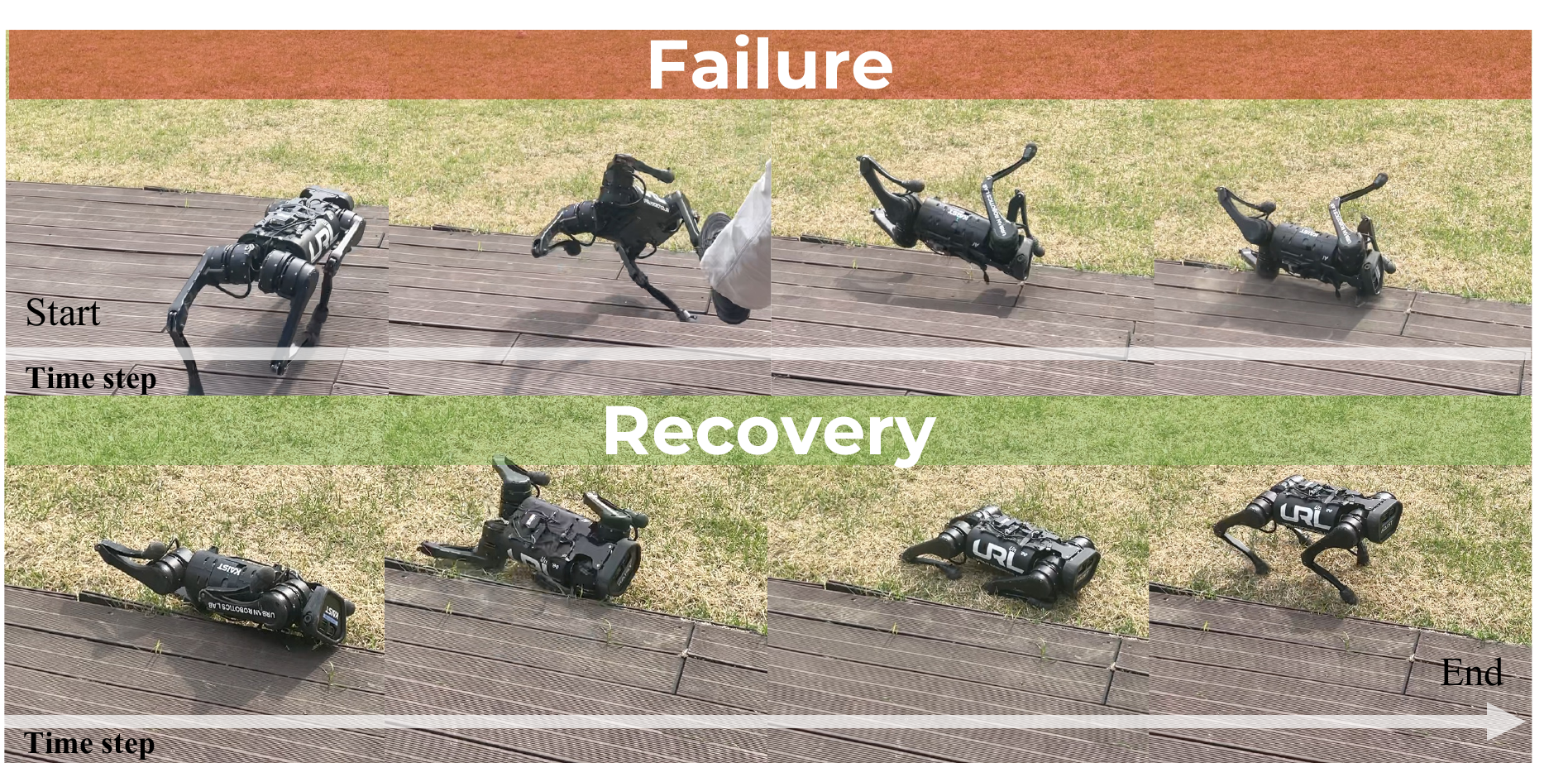}
	\end{subfigure}
	\captionsetup{font=footnotesize}
	\caption{Failure recovery scenario in quadrupedal locomotion.}
	\label{figure:mainfig}
    \vspace{-0.2cm}
\end{figure}

In this study, we introduce \textit{DreamRiser}, a robust all--terrain recovery motion control policy learning framework that incorporates an implicit perception of the surrounding terrain structure. By acquiring this capability, the recovery control policy can effectively restore the robot's pose to a stable standing position across diverse and unstructured terrains (Fig.~\ref{figure:mainfig}). Our approach builds upon DreamWaQ~\cite{nahrendra2023dreamwaq}, a locomotion policy learning framework that facilitates the implicit imagination of terrains. 

In summary, the contributions of this study are twofold:
\begin{enumerate}
    \item A recovery control policy framework that possesses adaptability to various terrain structures. This framework enables the robot to effectively recover its pose in different types of terrain.
    \item Comprehensive evaluations of our approach through simulations and real--world experiments that empirically demonstrates the effectiveness and robustness of our proposed recovery control policy.\footnote[1]{\label{note1}\url{https://sites.google.com/view/dreamriser}}
\end{enumerate}

The remainder of this paper is organized as follows. Section~\ref{section:DreamRiser} discusses our proposed method thoroughly. Section~\ref{section:experiments} presents the experimental setting, results, and an in-depth comparative analysis of the proposed and baseline methods. Finally, Section~\ref{section:conclusion} concludes this work and briefly discusses directions for future work.

\section{Methodology of DreamRiser}\label{section:DreamRiser}

\subsection{Preliminaries}
We model the environment as a partially observable Markov decision process (POMDP)~\cite{sutton2018reinforcement}, defined by the tuple $\mathcal{M}=(\mathcal{S},\mathcal{O},\mathcal{A},d_0,p,r,\gamma)$. The full state, partial observation, and action are continuous, and defined by $\mathcal{\textbf{s}\in\mathcal{S}}$, $\mathcal{\textbf{o}\in\mathcal{O}}$, and $\mathcal{\textbf{a}\in\mathcal{A}}$, respectively. The environment starts with an initial state distribution, $d_0(\textbf{s}_0)$; progresses with a state transition probability $p(\textbf{s}_{t+1}|\textbf{s}_{t},\textbf{a}_{t})$; and each transition is rewarded with a reward function, $r:\mathcal{S}\times\mathcal{A}\to\mathcal{R}$. The discount factor is defined by $\gamma\in[0,1)$. The temporal observation at time $t$ with the past $H$ measurements is defined as $\textbf{o}^H_t=\begin{bmatrix}\textbf{o}_t & \textbf{o}_{t-1} \dots \textbf{o}_{t-H} \end{bmatrix}^T$.  

\subsection{Task Formulation}\label{section:task_formulation}
The primary goal of a recovery task for a quadrupedal robot is to restore the robot's pose and joints to a state where it can walk normally using any locomotion controller.
The recovered state can be defined if two conditions are fulfilled: 1)~all the robot's feet are in contact with the ground, and 2)~the robot's base reached an upright pose. The first condition is the main distinction between DreamRiser's task formulation and related works~\cite{lee2019robust,smith2022legged}. Although achieving an upright base orientation is one of the main objectives, the robot's recovered pose may not be upright on some terrains that is bumpy and uneven, which leads to unstable recovered pose. Thus, training the robot to ensure stable foot contact with the terrain can help to stabilize its final recovery pose before entering locomotion mode.

\subsection{Terrain Imagination via Proprioception}\label{section:implicit}
To recover from various terrains, the robot needs to recognize the surrounding terrain's properties. One approach to achieve this is by incorporating a dedicated terrain mapping module~\cite{fankhauser2018probabilistic, miki2022elevation, hoeller2022neural}. However, in situations where the robot has flipped over or experienced a failure, exteroceptive mapping algorithms may not be applicable. In such scenarios, proprioception becomes the sole means for the robot to comprehend the terrain properties. By relying proprioceptive measurements, the robot can gain insights into the terrain features and adjust its recovery strategy accordingly.

Prior studies have demonstrated that terrain properties can be estimated using only proprioception~\cite{lee2020learning,kumar2021rma,nahrendra2023dreamwaq}. In this study, we incorporate the concept of implicit terrain imagination from DreamWaQ~\cite{nahrendra2023dreamwaq} to develop a robust recovery policy capable of effectively navigating diverse terrains. A key feature of DreamWaQ is its utilization of variational inference methods in the context-aided estimator network~(CENet) for predicting the latent properties of the surrounding terrains. This approach makes our policy more resilient to epistemic uncertainties present in real-world scenarios.

\subsubsection{Policy Network}
The policy, $\pi_{\phi}(\textbf{a}_t|\textbf{o}_t,\textbf{v}_t,\textbf{z}_t)$ is a neural network parameterized by $\phi$. The policy network infers an action $\textbf{a}_t\!\in\!\mathbb{R}^{12\times1}$, given a proprioceptive observation $\textbf{o}_t\!\in\!\mathbb{R}^{252\times1}$, body velocity $\textbf{v}_t\!\in\!\mathbb{R}^{3\times1}$, and latent state $\textbf{z}_t\!\in\!\mathbb{R}^{32\times1}$. $\textbf{o}_t$ consists of $\boldsymbol{\omega}_t$, $\textbf{g}_t$, $\textbf{c}_t$, $\boldsymbol{\theta}_t$, $\boldsymbol{\dot{\theta}}_t$, and $\textbf{a}_{t-1}$, which are the body angular velocity, gravity vector in the body frame, body velocity command, joint angle, joint angular velocity, and previous action, respectively. The CENet estimates $\textbf{v}_t$ and $\textbf{z}_t$ and trained jointly with the policy. 

\subsubsection{Value Network}
The value network is trained to estimate the state value, $V(\textbf{s}_t)$, given the the privileged observation, $\textbf{s}_t$,  which is defined as
\begin{equation}
  \textbf{s}_t=\begin{bmatrix}\textbf{o}_{t} & \textbf{v}_{t} & \textbf{d}_{t} & \textbf{h}_{t}\end{bmatrix}^T,
  \label{eqn:privileged_vector}
\end{equation}
where $\textbf{d}_{t}$ is the disturbance force applied randomly on the robot's body and $\textbf{h}_{t}$ is the height map scan of the robot's surroundings as an exteroceptive cue for the value network. 

\subsubsection{Action Space}
We use the target joint angle around the robot's self--righted pose, as the action space to facilitate learning and the target joint angle can be computed as
\begin{equation}
  \boldsymbol{\theta}_\text{target}= \boldsymbol{\theta}_\text{stand} +  \textbf{a}_t,
\end{equation}
where $\boldsymbol{\theta}_\text{target}$ and $\boldsymbol{\theta}_\text{stand}$ are the target joint angle and robot's default joint angle for self--righted, respectively. Each target joint angle is converted into torque using a proportional--derivative (PD) controller.

\subsection{Reward Function}
Our reward function consists of task and behavior objectives. The task objective is tracking an upright condition, inspired by the orientation tracking reward in~\cite{nahrendra2022retro} which is defined by aligning the gravity body frame's $z$-axis with the negative of the gravity vector. The behavior objectives are used for constraining aggressive motion during recovery by penalizing rapid motor  movement. The reward, $r_t(\textbf{s}_t,\textbf{a}_t)$, is defined as the weighted sum of all individual reward terms summarized in Table~\ref{table:reward_function}.

\begin{table}[t!]
\footnotesize
\centering
\captionsetup{font=footnotesize, singlelinecheck=false}
\caption{Reward function elements. $g_z$ is the $z$-axis component of the gravity vector projected to the robot's body frame. $c_\textrm{foot}$ is the foot contact state with values between $1$ (in contact) or $0$ (not in contact). $\textbf{a}_t$ is the policy's action at time $t$. $\boldsymbol{\dot{\theta}}$, $\boldsymbol{\ddot{\theta}}$, and $\boldsymbol{\tau}$ are joint velocity, acceleration, and torque, respectively.}
\label{table:reward_function}

\begin{center}
\begin{tabular}{lll}
\hline \hline
Reward           & Equation ($r_i$) & Weight ($w_i$) \\ \hline 
Base uprightness & $1-g_z$ & $1.0$ \\
Foot contact & $c_\textrm{foot}$ & $1.0$ \\
Joint accelerations & $\boldsymbol{\ddot{\theta}}^2$ & $-10^{-6}$ \\
Joint power & $\abs{\boldsymbol{\tau}}|\boldsymbol{\dot{\theta}}|$ & $-10^{-5}$ \\
Action rate & $(\textbf{a}_t - \textbf{a}_{t-1})^2$ & $-0.05$ \\
\hline \hline
\end{tabular}
\end{center}
\vspace{-0.4cm}
\end{table}

\subsection{Terrain Curriculum}~\label{section:curriculum}
We employ a simple training curriculum to emphasize the learned policy's adaptability beyond its training distribution. The robot was dropped from various poses onto discrete terrains at the beginning of each episode. These discrete terrains were characterized by an escalating range of terrain height spanning from $[0, 0.1]$ up to $[0, 1.0]$ with ten different difficulty levels. In each increasing level, the maximum terrain height is increased by $10$ $\textrm{cm}$.

\section{Experiments}~\label{section:experiments}
\vspace{-0.3cm}

\subsection{Training in Simulation}
We trained the policy in simulation using the Isaac Gym simulator~\cite{makoviychuk2021isaac} and legged robot gym library~\cite{rudin2022learning}. We parallelized the training process with $4,\!096$ domain--randomized agents with randomized parameters, as reported in Table~\ref{table:domain_randomization}. We employed the proximal policy optimization (PPO) algorithm~\cite{schulman2017proximal} to update the policy and value networks. We set the clipping range, generalized advantage estimation factor, and discount factor as $0.2$, $0.95$, and $0.99$, respectively. All networks were optimized with a learning rate of $10^{-3}$ using the Adam optimizer~\cite{kingma2014adam}. The training was performed on a desktop PC with an Intel Core i7-8700 CPU @ 3.20 GHz, 32 GB RAM, and an NVIDIA RTX 3080Ti GPU. 

\subsection{Robot Specification}\label{section:robot_specification}
For our experiments, we employed the Unitree A1~\cite{unitreea1} and Unitree Go1~\cite{unitreego1} robots to evaluate the performance of our learning-based recovery controller, both with and without payloads. The policy execution was synchronized with the CENet at a frequency of 50 Hz. To track the desired joint angles, a PD controller was employed with proportional and derivative gains set to $K_p\!=\!28$ and $K_d\!=\!0.7$, respectively, operating at a frequency of 200 Hz. For real-world deployment, all neural networks were implemented on the onboard NVIDIA Jetson NX utilizing Torch JIT optimization.

\begin{table}[t!]
\footnotesize
\centering
\captionsetup{font=footnotesize, justification=centering}
\caption{Domain randomization ranges applied in the simulation.}
\label{table:domain_randomization}
\begin{center}
\scriptsize
\begin{tabular}{lccc}
\hline\hline
Parameter                        & Randomization range & Units \\\hline
Payload                     & $[-1, 2]$ & $\mathrm{~kg}$     \\ 
$K_p$ factor   & $[0.9, 1.1]$ & $\mathrm{~Nm/rad}$    \\ 
$K_d$ factor        & $[0.9, 1.1]$ & $\mathrm{~Nms/rad}$   \\ 
Motor strength factor        & $[0.9, 1.1]$ & $\mathrm{~Nm}$   \\ 
Center of mass shift        & $[-50,50]$ & $\mathrm{~mm}$    \\
\hline\hline
\end{tabular}
\end{center}
\vspace{-0.2cm}
\end{table}

\subsection{Success Rates}
\begin{figure}[t!]
	\centering 
	\begin{subfigure}[]{0.23\textwidth}
		\includegraphics[width=1.0\textwidth]{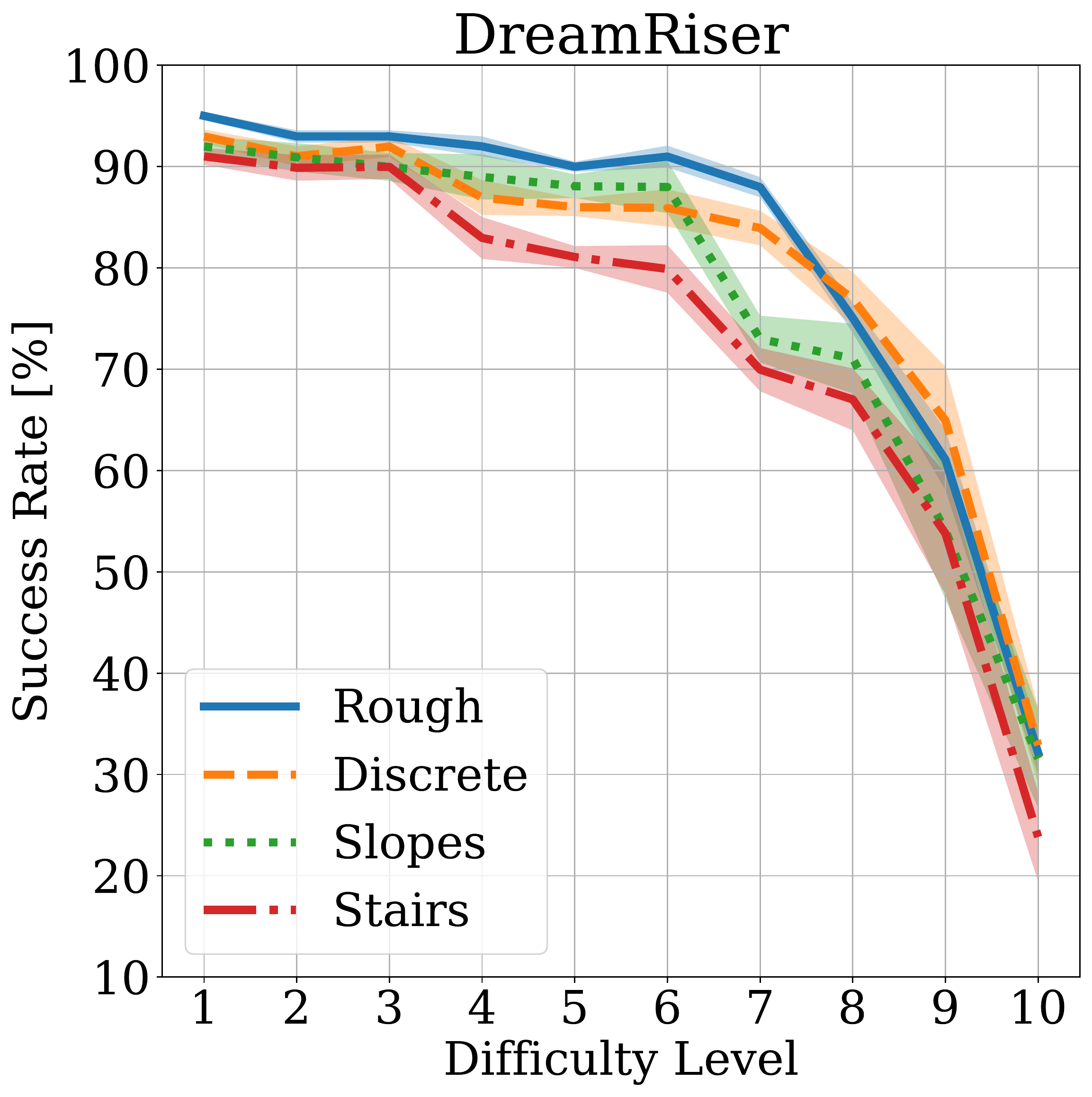}
	\end{subfigure}
    \begin{subfigure}[]{0.23\textwidth}
		\includegraphics[width=1.0\textwidth]{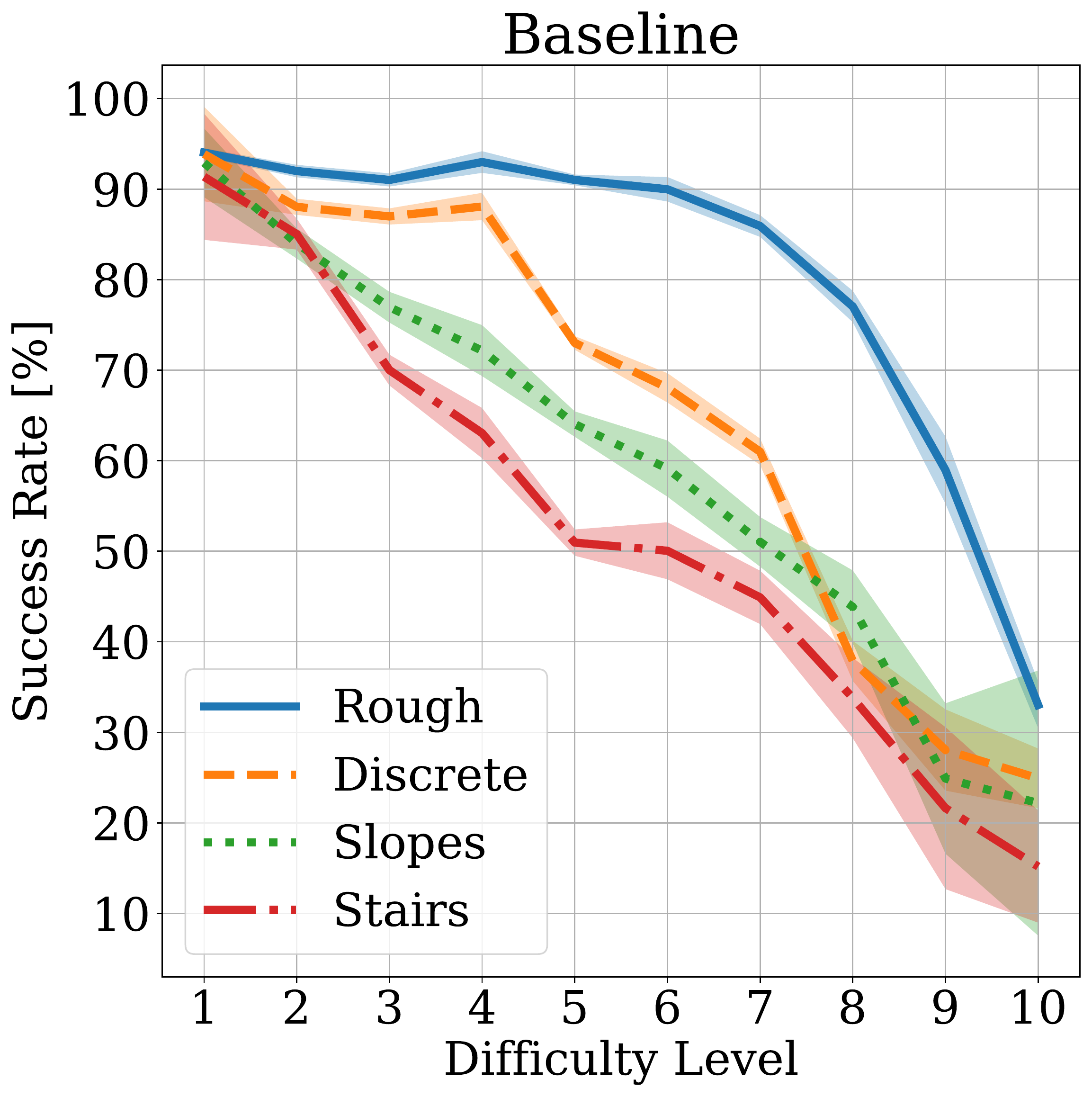}
	\end{subfigure}
	\captionsetup{font=footnotesize}
	\caption{Recovery success rate on different environments. The lines indicate mean of the success rates, while shaded regions indicate the standard deviation of the success rates from ten random seeds.}
	\label{figure:success_rate}
\end{figure}

To quantitatively assess the robustness of the learned policies, we compared the DreamRiser policy with a baseline policy~\cite{lee2019robust} that was trained using vanilla end-to-end RL without CENet and asymmetric actor--critic. We experimented using simulated robots in environments with different levels of difficulty. The difficulty levels were discretized into ten distinct levels by discretizing the parameter range, gradually increasing in complexity as the level progresses as follows:
\begin{enumerate}
    \item \textbf{Rough}: Rough terrain with increasing level of terrain noise within $[-0.5,0.5]$ $\textrm{m}$.
    \item \textbf{Discrete}: Discretized blocks spawned randomly on a flat terrain with box size within $[-0.5,0.5]$ $\textrm{m}$.
    \item \textbf{Slopes}: Slope with increasing levels of angle between $[10.0, 30.0]$ $\textrm{deg}$.
    \item \textbf{Stairs}: Fixed--width stairs with increasing levels of angle between $[10.0, 30.0]$ $\textrm{deg}$.
\end{enumerate}
In this quantitative evaluation, $1,\!000$ robots were deployed in the same environment. A successful recovery is defined as the robot achieving a stable upright pose within five seconds. The number of robots that successfully recovered was recorded to measure the success rates. The results shown in Fig.~\ref{figure:success_rate} highlight that DreamRiser's policy is more robust and enables the robot to recover in a wide variety of terrains.

\subsection{Sim-to-Real Transfer}
\begin{figure*}[t!]
	\centering 
	\begin{subfigure}[b]{0.93\textwidth}
		\includegraphics[width=1.0\textwidth]{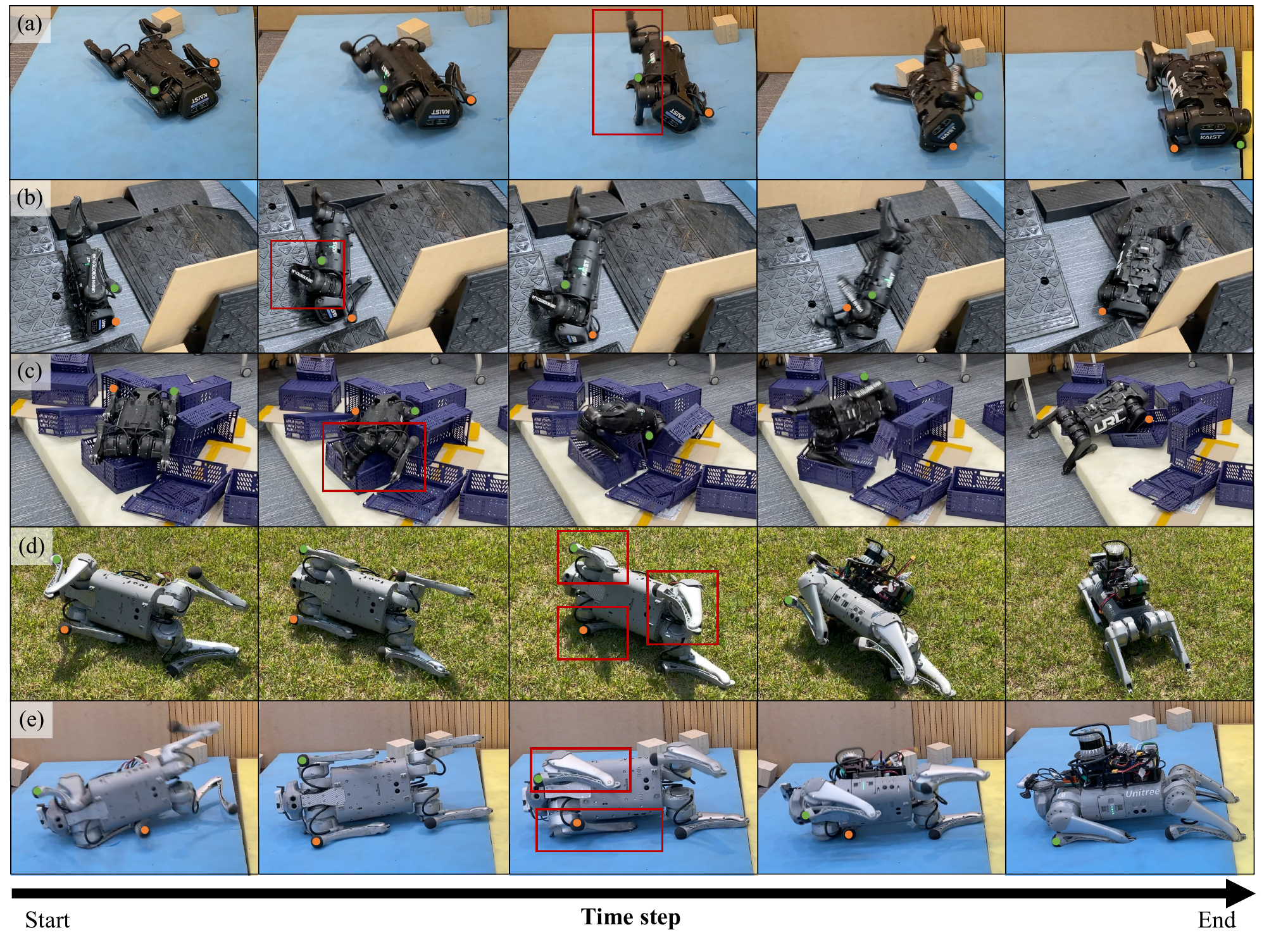}
	\end{subfigure}
	\captionsetup{font=footnotesize}
	\caption{Recovery motion using the policy learned with DreamRiser\textsuperscript{\ref{note1}}. Red boxes in the snapshots highlight the adaptive recovery motion. The recovery controller enabled the robot to recover its pose in various terrains such as (a)~sponge, (b)~irregular bumps, (c)~piles of boxes, and (d)-(e)~with payloads on top of the robot. The recovery motion is not limited to a single predefined motion but instead allows for adaptive and dynamic adjustments by quickly assessing the terrain properties.}
	\label{figure:simtoreal_recovery}
 \vspace{-0.4cm}
\end{figure*}

To further assess the robustness of the recovery control policy, we conducted real-world tests under various settings as shown in Fig.~\ref{figure:mainfig}. By subjecting the recovery control policy to such diverse conditions, we validated its ability to recover the robot's pose reliably across a range of challenging terrains. These experiments provide insights into the policy's performance and adaptability to different terrain types and external load conditions without directly measuring the terrain properties. The adaptive recovery motions are highlighted by red boxes on the snapshots in Fig.~\ref{figure:simtoreal_recovery}. We will consistently use the same color to indicate \textcolor{greenfoot}{\textbf{left}} and \textcolor{orangefoot}{\textbf{right}} hemispheres of the robot's body throughout the paper.

In Figs.~\ref{figure:simtoreal_recovery}(a) and \ref{figure:simtoreal_recovery}(b), the robot demonstrates a deep hip abduction on its \textcolor{greenfoot}{\textbf{front left leg}} to lift its body and the \textcolor{greenfoot}{\textbf{rear left leg}} performs a swing motion to generate a rolling moment. Meanwhile the \textcolor{orangefoot}{\textbf{right legs}} are used to support the whole body during the rolling motion. In Fig.~\ref{figure:simtoreal_recovery}(c), the robot is placed on top of boxes. It firstly folds all of its legs to search for any available surface to initiate the rolling motion. Afterwards, it swiftly swings its legs to generate moments that ease the rolling of its body. The robot is equipped with additional payloads in Figs.~\ref{figure:simtoreal_recovery}(d) and \ref{figure:simtoreal_recovery}(e). It performs minimum leg motions to avoid collision with the payload. It firstly moves the legs that are in contact with the surface to find a stable support and swings the other legs to generate a momentum to roll over its body.


\subsection{Embedding Analysis}
\begin{figure}[t!]
	\centering 
	\begin{subfigure}[b]{0.48\textwidth}
		\includegraphics[width=1.0\textwidth]{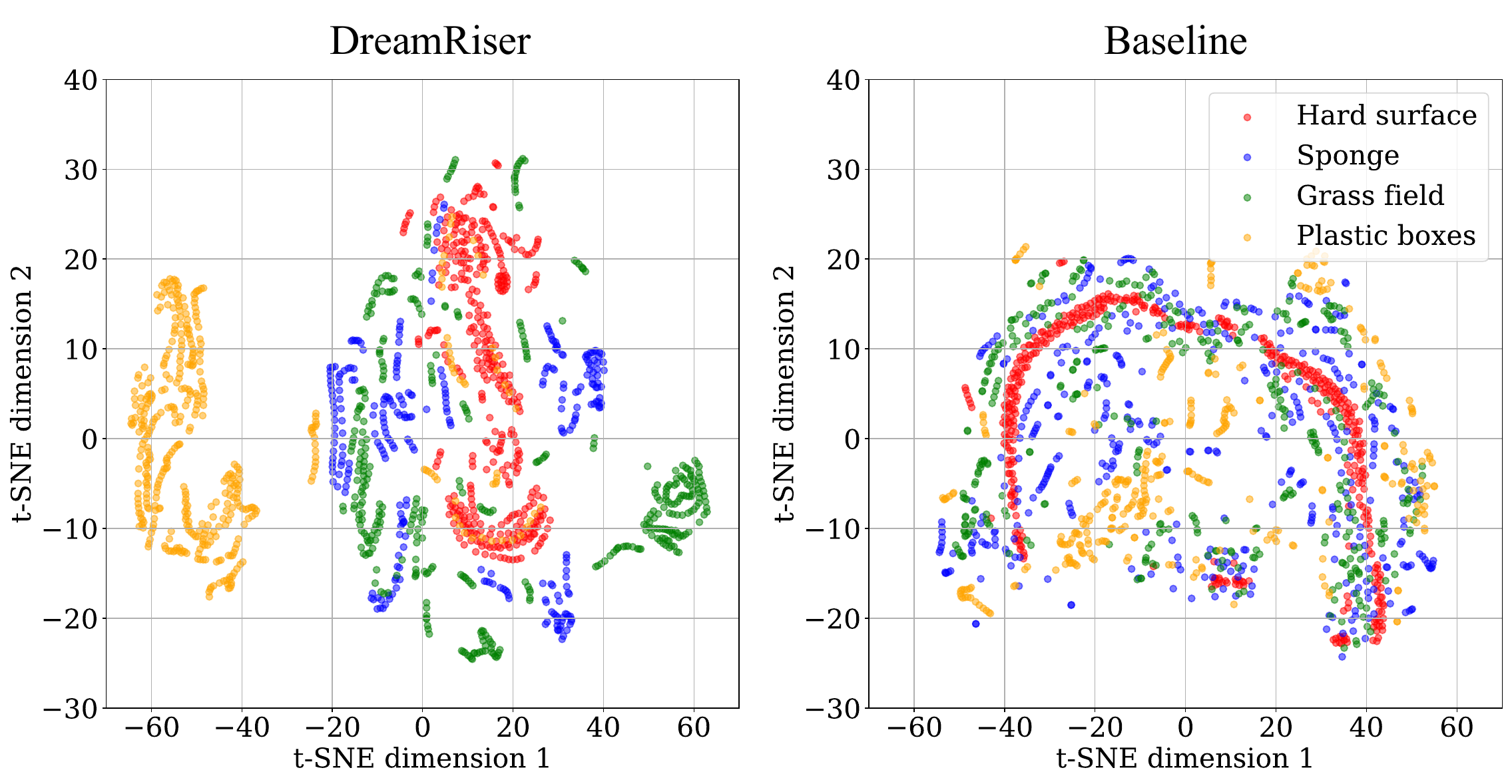}
	\end{subfigure}
	\captionsetup{font=footnotesize}
	\caption{Two--dimensional t-SNE plot for qualitative embedding analysis of our proposed DreamRiser and baseline approach. Latent embeddings from DreamRiser have better disentanglement than the baseline, as shown by more distinct cluster points in the plot.}
	\label{figure:embedding_analysis}
 \vspace{-0.4cm}
\end{figure}

To gain further insights, we recorded and then visualized the latent states inferred by the CENet when the robot entered the recovery mode. After recording the latent states, we performed a t-distributed stochastic neighbor embedding (t-SNE) dimensionality reduction on these latent states and visualize them in a 2D space as depicted in Fig.~\ref{figure:embedding_analysis}.

We compared the latent embeddings inferred by DreamRiser's CENet and the baseline to highlight the significance of terrain imagination in enhancing the performance of the recovery controller. The latent states inferred by the CENet have a higher degree of disentanglement, indicated by a more distinct clustering in the t-SNE plot. Disentangled latent representation plays a vital role in enabling the policy to quickly distinguish between different terrain properties and adapt its recovery motion accordingly as shown in Fig.~\ref{figure:simtoreal_recovery}. For instance, the latent embeddings from DreamRiser on the pile of boxes experiments are located quite distant from the other states, which explains why the recovery motion in Fig.~\ref{figure:simtoreal_recovery}(c) is significantly different from the other scenarios.


\section{Conclusion}~\label{section:conclusion}
In this study, we present a robust recovery control policy learning framework to facilitate robust pose recovery of quadrupedal robots across diverse terrain conditions. We conducted thorough evaluations, both in simulation and real-world environments, to demonstrate the effectiveness and robustness of the learned recovery policy. The results showcase successful pose recovery of quadrupedal robots, corroborating the practical applicability of our approach.

As a future work, we intend to integrate recovery policy learning with a fall detector and locomotion policy learning. This integration seeks to enable the locomotion policy to learn a unified locomotion and recovery policy that can quickly respond to potential failures. This enhanced capability will contribute to the overall autonomy and reliability of quadrupedal robots in challenging environments.

\bibliographystyle{IEEEtran}
\bibliography{./main,./IEEEabrv}

\end{document}